\documentclass[10pt,twocolumn,letterpaper]{article}

\usepackage{cvpr}
\usepackage{times}
\usepackage{epsfig}
\usepackage{graphicx}
\usepackage{amsmath}
\usepackage{amssymb}
\usepackage{caption}
\usepackage{url}

\usepackage[breaklinks=true,bookmarks=false]{hyperref}

\cvprfinalcopy 

\def\cvprPaperID{108} 

\ifcvprfinal\fi

\begin{document}

\title{WiCV 2020: The Seventh Women In Computer Vision Workshop}

 \author{
 Hazel Doughty$^1$, Nour Karessli$^2$, Kathryn Leonard$^3$, Boyi Li$^4$,\\ Carianne Martinez$^5$, Azadeh Mobasher$^6$, Arsha Nagrani$^7$, Srishti Yadav$^8$\\\\
$^1$University of Bristol, $^2$Zalando SE, $^3$Occidental College,  $^4$Cornell University,\\ $^5$Sandia National Laboratories, $^6$Microsoft, $^7$University of Oxford, $^8$Simon Fraser University\\

 \tt\small wicv2020-organizers@googlegroups.com 
}



\maketitle

\begin{abstract}
\thispagestyle{empty}
In this paper we present the details of Women in Computer Vision Workshop - WiCV 2020, organized in alongside virtual CVPR 2020. This event aims at encouraging the women researchers in the field of computer vision. It provides a voice to a minority (female) group in computer vision community and focuses on increasingly the visibility of these researchers, both in academia and industry. WiCV believes that such an event can play an important role in lowering the gender imbalance in the field of computer vision. WiCV is organized each year where it provides a.) opportunity for collaboration with between researchers b.) mentorship to female junior researchers  c.) financial support to presenters to overcome monetary burden and d.) large and diverse choice of role models, who can serve as examples to younger researchers at the beginning of their careers. In this paper, we present a report on the workshop program, trends over the past years, a summary of statistics regarding presenters, attendees, and sponsorship for the current workshop.

\end{abstract}

\section{Introduction}
While excellent progress has been made in a wide variety of computer vision research areas in recent years, similar progress has not been made in the increase of diversity in the field and the inclusion of all members of the computer vision community. Despite the rapid expansion of our field, female research still only account for a small percentage of the researchers in both academia and industry. Due to this, many female computer vision researchers can feel isolated in workspaces which remain unbalanced due to the lack of inclusion.

The Women in Computer Vision workshop is a gathering for both women and men working in computer vision. It aims to appeal to researchers at all levels, including established researchers in both industry and academia (e.g. faculty or postdocs), graduate students pursuing a Masters or PhD, as well as undergraduates interested in research.  This aims to raise the profile and visibility of female computer vision researchers each of these levels, seeking to reach women from diverse backgrounds at universities and industry located all over the world. 

There are three key objectives of the WiCV workshop.
The first to increase the WiCV network and promote interactions between members of the WiCV network, so that female students may learn from professionals who are able to share career advice and past experiences. A mentoring banquet is run alongside the workshop. This provides a casual environment where both junior and senior women in computer vision can meet, exchange ideas and even form mentoring or research relationships.

The workshop's second objective is to raise visibility of women in computer vision. This is done at both the junior and senior levels. Several senior researchers are invited to give high quality keynote talks on their research, while junior researchers are invited to submit recently published or ongoing works with many of these being selected for oral or poster presentation through a peer review process. This allows junior female researchers to gain experience presenting their work in a professional yet supportive setting. We strive for diversity in both research topics and presenters' background. The workshop also includes a panel, where the topics of inclusion and diversity can be discussed between female and male colleagues.

Finally, the third objective is to offer junior female researchers the opportunity to attend a major computer vision conference which they otherwise may not have the means to attend. This is done through travel grants awarded to junior researchers who present their work in the workshop via a poster session. These travel grants allow the presenters to not only attend the WiCV workshop, but also the rest of the CVPR conference. 

\section{Workshop Program}
\label{program}

The workshop program consisted of 4 keynotes, 4 oral presentations, 37 poster presentations, a panel discussion, and a mentoring session. As with previous years, our keynote speakers were selected to have diversity among topic, background, whether they work in academia or industry, as well as their seniority. It crucial to provide a diverse set of speakers so that junior researchers have many different potential role models who they can related to in order to help them envision their own career paths.

The workshop schedule was as follows:
\begin{itemize}
  \setlength\itemsep{-0.05em}
\item Introduction
\item Invited Talk 1: Chelsea Finn (Stanford University), \textit{Generalization in Visuomotor Learning}
\item Oral Session
\begin{itemize}
\item Medhini Narasimhan (UC Berkeley), \textit{Seeing the Un-Scene: Learning Amodal Semantic Maps for Room Navigation}
\item Safa Messaoud (UIUC), \textit{Can We Learn Heuristics For Graphical Model Inference Using Reinforcement Learning}
\item Rangel Daroya (University of the Philippines), \textit{REIN: Flexible Mesh Generation from Point Clouds}
\item Sayna Ebrahimi (UC Berkeley), \textit{Variational Adversarial Active Learning}
\end{itemize}
\item Invited Talk 2: Tali Dekel (Google AI Research), \textit{Learning to Retime People in Videos}
\item Poster Session (repeated 12 hours later)
\item Invited Talk 3: Georgia Gkioxari (Facebook AI Research), \textit{Beyond 2D Visual Recognition}
\item Invited Talk 4: Kavita Bala (Cornell University \& Facebook), \textit{Visual Understanding at Global Scale}
\item Panel Session
\item Closing Remarks
\item Mentoring Session 
\begin{itemize}
\item Speaker: Kristen Grauman (UTAustin)
\end{itemize}
\end{itemize}

\subsection{Virtual Setting}
This year since the organization has been slightly modified as CVPR 2020 was held remotely. We made sure to make the virtual WiCV workshop as engaging and interactive as possible. We used Zoom as our virtual platform, which provided us the opportunity to conduct talks and panel discussion seamlessly. Breakout sessions were used for the poster session, allowing easy display of a poster or video and live discussion amongst presenters and attendees. The workshop was set up to have two poster presentation sessions with a 12 hours difference between them to cope with differing timezones. This maximised the amount of people able to attend the workshop and see the accepted works. All the presented keynotes and talks have been uploaded to our website to allow asynchronous viewing after the workshop was over.

\section{Workshop Statistics}

Originally, the first workshop for WiCV was held in conjunction with CVPR 2015. Since then, the participation rate and number quality of submissions to WiCV have been steadily increasing. 
Following the examples from the editions held in 2018 and 2019\cite{Amerini19,Akata18,Demir18}, we were encouraged to collect the top quality submissions into workshop proceedings. By providing oral and poster presenters with the opportunity to publish their work in the conference's proceedings, we believe that the visibility of female researchers will be further increased.
This year, the workshop was held as a half day virtual gathering over Zoom webinar. Senior and junior researchers were invited to present their work, and poster presentations are included as already described in the previous Section \ref{program}.\\

The organizers for this year WiCV workshop are working in both academia and industry from various institutions located in six different timezones, with up to twelve and half hours difference. Their miscellaneous backgrounds and research areas have pledged the organizing committee a diverse perspective. Their research interests in computer vision and machine learning include video understanding, computational geometry, representation learning, uncertainty modelling, optimization, multi-modal and semi-supervised learning in different industrial application areas such as healthcare,  and fashion.

\begin{figure}[h]
    \centering
    \includegraphics[width=1\linewidth]{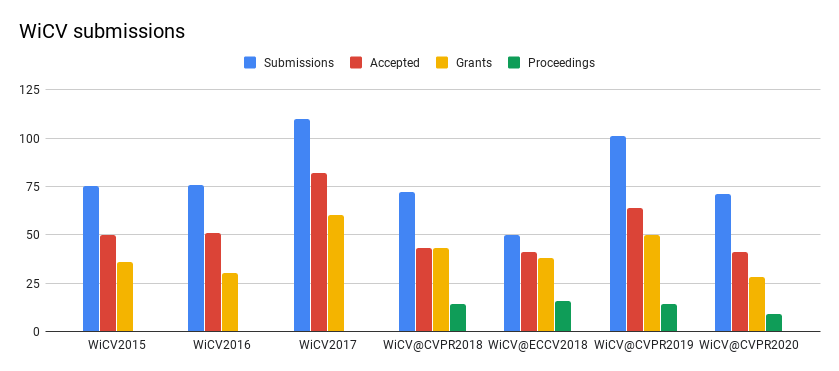}
    \captionof{figure}{\textbf{WiCV Submissions.} The number of submissions over the past years of WiCV.}
    \label{fig:sub}
\end{figure}

This year we had 71 high quality submissions from a wide range of topics and institutions. This is slightly reduced from previous year due to the global pandemic. It is on par WiCV@CVPR18, the previous CVPR edition which also had an ECCV\cite{Akata18} edition in the same year. The most popular topics were object recognition and deep learning and convolutional neural networks followed by video understanding, images and language and vision for robotics. 
Over all submissions, around 6\% were selected to be presented as oral talks and 52\% were selected to be presented as posters. Within the accepted submissions, 9 were 4-8 page papers selected to be included in the workshop's proceedings. The comparison with previous years is presented in Figure~\ref{fig:sub}. With the great effort of an interdisciplinary program committee consisting of 61 reviewers, the submitted papers were evaluated and received valuable feedback.

Although this edition was exceptionally held remotely, we have persisted WiCV tradition of last year's workshops \cite{Amerini19,Akata18,Demir18} in providing grants to help the author's of accepted submissions participate in the workshop. The grants covered the conference workshop registration fees for all the authors of accepted submissions who requested funding. In addition, we have introduced a new raffle for the workshop participants. The prize is Titan RTX GPU provided by Nvidia.

\begin{figure}
    \centering
    \includegraphics[width=1\linewidth]{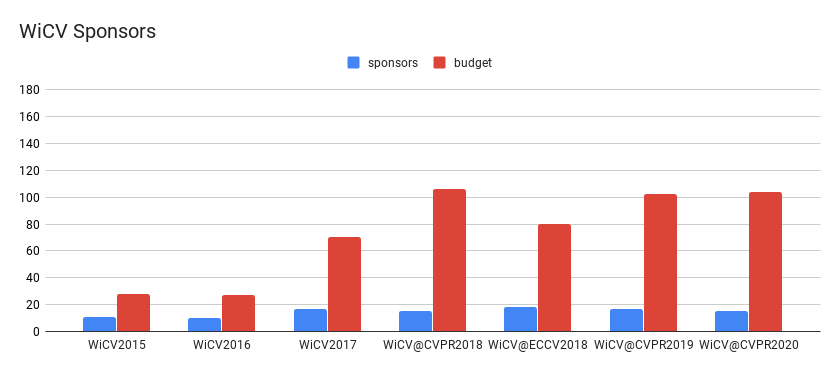}
    \captionof{figure}{\textbf{WiCV Sponsors.} The number of sponsors and the amount of sponsorships for WiCV. The amount is expressed in US dollar (USD).}
    \label{fig:spo}
\end{figure}

The total amount of sponsorship this year is \$104 USD and a GPU with 15 sponsors, reaching a very good target. In Figure~\ref{fig:spo} you can find the details respect to the past years. The majority of this sponsorship was spent on registration stipends with some of the sponsorship also going towards covering the cost of the Zoom webinar.

\section{Conclusions}
WiCV at CVPR 2020 has continued to be a valuable opportunity for presenters, participants and organizers in providing a platform to bring the community together. It continues to overcome the existing issue of gender balance prevailing around us and we hope that it has played an important part to making the community even stronger. Being a virtual workshop and first of its kind in the history of WiCV, it provided an opportunity for people to connect from all over the world from their personal comforts. With distance and commute no more an issue, virtual workshop brought the community even closer. With high number of paper submissions and even higher number of attendees, we foresee that the workshop will continue the marked path of previous years and foster stronger community building with increased visibility, providing support, and encourage inclusively for all the female researchers in academy and in industry.

\section{Acknowledgments}
First of all we would like to thank our sponsors. We are very grateful to our other Platinum sponsors: Toyota Research Institute, Microsoft, Apple, Amazon, Facebook, Netflix, Google Research and IBM Research. We would also like to thank our Gold sponsor: Uber; Silver Sponsors: Disney Research, Intel AI and the Allen Institute for AI; Bronze sponsors: Lyft and Waymo, as well as Nvidia for providing us with the Titan RTX GPU raffle prize. We would also would like to thank Occidental College as our fiscal sponsor, which donated employee knowledge and time to process our sponsorships and travel awards. We would also like to thank and acknowledge the organizers of WiCV at CVPR 2019, without the information flow and support from the previous WiCV organizers, this WiCV would not have been possible. We would like to also acknowledge CVPR 2020 Workshop Chairs, Tali Dekel and Tan Hassner and the Publications Chairs Eric Mortensen and Margaux Masson for answering all questions concerns timely. Huge thanks also go to Terry Boult who made a mammoth effort to make a virtual CVPR possible. Finally, we would like to acknowledge the time and efforts of our program committee, authors, reviewers, submitters, and our prospective participants for being part of WiCV network community.

\section{Contact}
\noindent \textbf{Website}: \url{http://sites.google.com/view/wicvworkshop-cvpr2020/}\\
\textbf{E-mail}: wicv2020-organizers@googlegroups.com\\
\textbf{Facebook}: \url{https://www.facebook.com/WomenInComputerVision/}\\
\textbf{Twitter}: \url{https://twitter.com/wicvworkshop}\\
\textbf{Google group}: women-in-computer-vision@googlegroups.com \\

{\small
\bibliographystyle{ieee}
\bibliography{egbib}
}

\end{document}